\documentclass[iicol,sn-basic]{sn-jnl}
\usepackage{graphicx}%
\usepackage{multirow}%
\usepackage{amsmath,amssymb,amsfonts}%
\usepackage{amsthm}%
\usepackage{mathrsfs}%
\usepackage[title]{appendix}%
\usepackage{xcolor}%
\usepackage{textcomp}%
\usepackage{manyfoot}%
\usepackage{booktabs}%
\usepackage{algorithm}%
\usepackage{algorithmicx}%
\usepackage{algpseudocode}%
\usepackage{listings}%
\usepackage{makecell}
\theoremstyle{thmstyleone}%
\theoremstyle{thmstyletwo}%

\theoremstyle{thmstylethree}%

\begin{document}

\title[Article Title]{CLIP-Powered TASS: Target-Aware Single-Stream Network for Audio-Visual Question Answering}

%%=============================================================%%
%% GivenName	-> \fnm{Joergen W.}
%% Particle	-> \spfx{van der} -> surname prefix
%% FamilyName	-> \sur{Ploeg}
%% Suffix	-> \sfx{IV}
%% \author*[1,2]{\fnm{Joergen W.} \spfx{van der} \sur{Ploeg} 
%%  \sfx{IV}}\email{iauthor@gmail.com}
%%=============================================================%%

\author[1]{\fnm{Yuanyuan} \sur{Jiang}}\email{jyy@bupt.edu.cn}

\author*[1]{\fnm{Jianqin} \sur{Yin}}\email{jqyin@bupt.edu.cn}

\affil*[1]{\orgdiv{School of Artificial Intelligence}, \orgname{Beijing University of Posts and Telecommunications}, \orgaddress{\street{Xitucheng Road 10}, \state{Haidian District}, \city{Beijing}, \postcode{100876}, \country{PR China}}}

%%==================================%%
%% Sample for unstructured abstract %%
%%==================================%%

\abstract{While vision-language pretrained models (VLMs) excel in various multimodal understanding tasks, their potential in fine-grained audio-visual reasoning, particularly for audio-visual question answering (AVQA), remains largely unexplored. AVQA presents specific challenges for VLMs due to the requirement of visual understanding at the region level and seamless integration with audio modality. Previous VLM-based AVQA methods merely used CLIP as a feature encoder but underutilized its knowledge, and mistreated audio and video as separate entities in a dual-stream framework as most AVQA methods. This paper proposes a new CLIP-powered target-aware single-stream (TASS) network for AVQA using the image-text matching knowledge of the pretrained model through the audio-visual matching characteristic of nature. It consists of two key components: the target-aware spatial grounding module (TSG+) and the single-stream joint temporal grounding module (JTG). Specifically, we propose a TSG+ module to transfer the image-text matching knowledge from CLIP models to our region-text matching process without corresponding ground-truth labels. Moreover, unlike previous separate dual-stream networks that still required an additional audio-visual fusion module, JTG unifies audio-visual fusion and question-aware temporal grounding in a simplified single-stream architecture. It treats audio and video as a cohesive entity and further extends the pretrained image-text knowledge to audio-text matching by preserving their temporal correlation with our proposed cross-modal synchrony (CMS) loss. Extensive experiments conducted on the MUSIC-AVQA benchmark verified the effectiveness of our proposed method over existing state-of-the-art methods. }

\keywords{Audio-visual question answering, Audio-visual learning, Scene understanding, Spatial-temporal reasoning, Pretrained knowledge}

%%\pacs[JEL Classification]{D8, H51}

%%\pacs[MSC Classification]{35A01, 65L10, 65L12, 65L20, 65L70}

\maketitle

\section{Introduction}\label{sec1}

\begin{figure}[t]
\centering
\includegraphics[width=7.6cm]{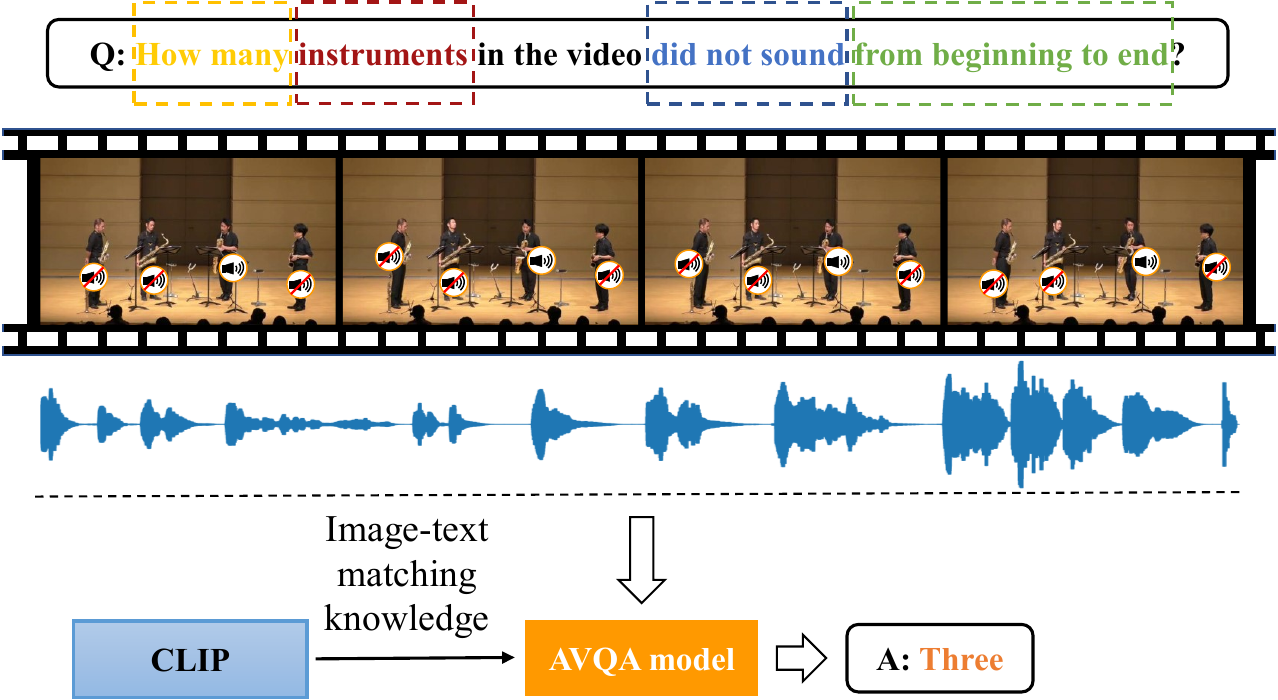}\\
\caption{An illustration of AVQA. Our proposed TASS-Net leverages the prior image-text matching knowledge from the pretrained model and transfers it to the AVQA model for spatio-temporal reasoning. The question is centered around the ``instruments'' (\textit{i.e.}, the target) and is broken down into ``how many'', ``did not sound'', and ``from beginning to end'' in terms of visual space, audio, and temporality, respectively. 
To answer this may entail a significant time investment for a human viewer, but an AI system with effective audio-visual scene parsing and spatio-temporal reasoning capabilities can achieve it promptly.}\label{fig1} 
\end{figure}

Audio-visual question answering (AVQA) has received considerable attention due to its potential applications in many real-world scenarios. It provides avenues to integrate multimodal information to achieve fine-grained spatio-temporal reasoning ability as humans \cite{MUSIC-AVQA}. As Fig. \ref{fig1} shows, AVQA aims to answer questions regarding visual objects, sound patterns, and their spatio-temporal associations. While large-scale vision-language pretrained models (VLMs), \textit{e.g.}, CLIP \cite{clip}, excel in various multimodal understanding tasks for their image-text matching capability, many studies \cite{rao2022denseclip,zhou2022extract,StructureCLIP} have revealed that vanilla CLIP has limited ability to understand finer-grained semantics between text and images. Therefore, employing CLIP as a feature extractor without fully leveraging its pretrained knowledge usually results in suboptimal performance in AVQA \cite{li2023progressive}.
% , which requires visual understanding at the region level rather than at the image level. 
Moreover, AVQA presents its specific challenges. Firstly, it involves effectively parsing audio and visual information related to the question in each timestamp, especially when there is ambient noise or similar categories. Secondly, it requires fusing the question-relevant audio-visual features across all timestamps while maintaining their temporal correlation in a multimedia video.

Current works \cite{jiang2023avel,tian2020unifiedavvp,wu2019dual} in the audio-visual scene understanding community address the first challenge by targetless parsing of audio-visual scenes.
% They \cite{xuan2020cross,wu2019dual,mercea2022avca} obtain untargeted sound-related visual regions by designing attention schemes performing on audio-to-visual.  
% For example, as illustrated in Figure \ref{fig1}, our focus lies solely on the subject of inquiry, \textit{i.e.}, \textit{instruments}, disregarding the singing person or ambient sound. 
Most AVQA approaches \cite{MUSIC-AVQA,yun2021pano,lao2023coca,LAVISH_CVPR2023}, inherited from the community, rely on aligning all audio-visual elements in the video to answer a question. However, the understanding of audio-visual scenes in AVQA is question-oriented. These approaches inevitably result in much irrelevant information and noise. \cite{jiang2023avqa} and \cite{li2023progressive} introduced the explicit semantics from the question into each timestamp, consequently improving performance. However, there is no natural alignment between visual and text as there is between visual and audio, thus impacting the overall model performance of \cite{jiang2023avqa}. \cite{li2023progressive}, on the other hand, although utilizing CLIP as the feature extractor for image-text alignment, did not make full use of its pretrained knowledge. 
% while only coarse-grained Q\&A labels are available. 
As most CLIP-powered approaches \cite{rao2022denseclip,zhou2022extract,StructureCLIP} necessitate fine-grained semantic ground truth labels to transfer image-level knowledge to a finer-grained level, \textit{e.g.}, pixel-level, AVQA only has coarse-grained Q\&A labels for videos but require fine-grained region-level visual understanding to achieve spatial reasoning. This inherent challenge limits the application of the CLIP model to AVQA.

\begin{figure}[t]
\centering
\includegraphics[width=7cm]{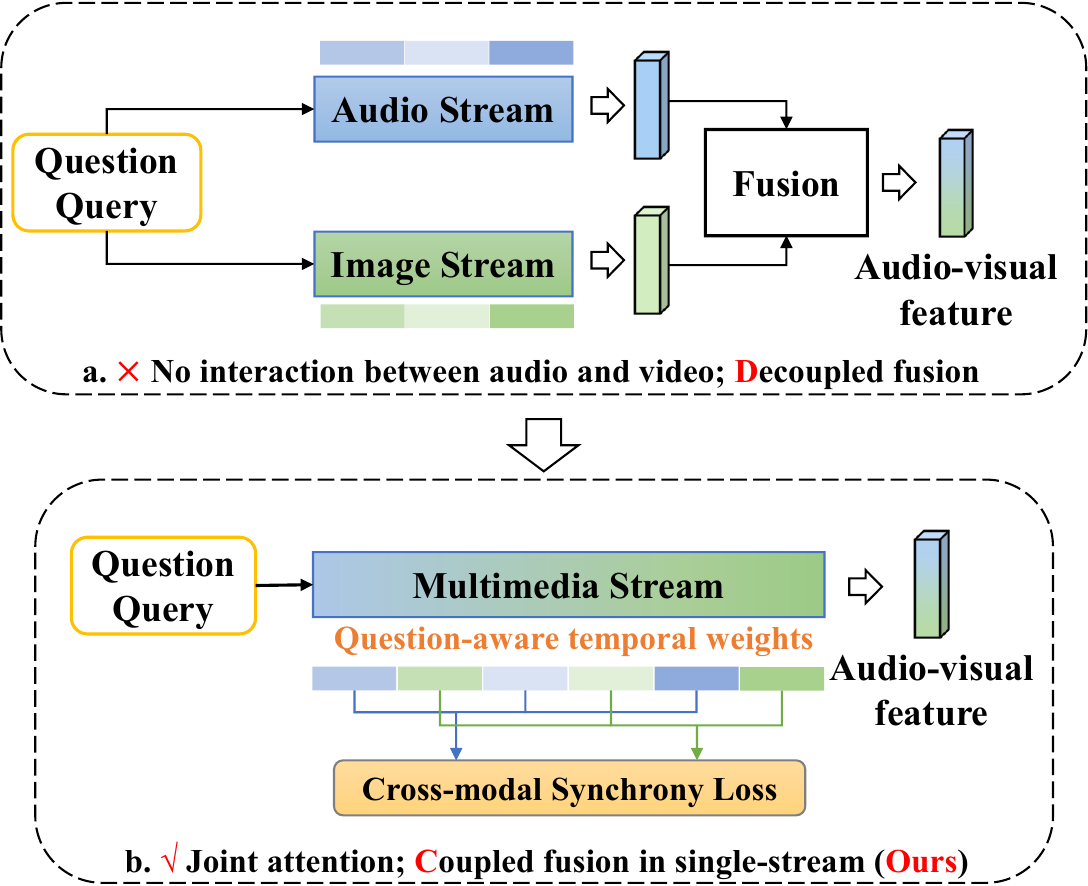}\\
\caption{Comparison of different question-aware temporal grounding. (a.) The traditional approach usually adopts a dual-stream network that treats audio and video as separate entities. (b.) Our proposed single-stream architecture treats audio and video as a whole and ensures the correlation between them, incorporating the two processes of temporal grounding and fusion.}\label{fig2} 
\end{figure}

As for the second challenge in AVQA, existing methods \cite{MUSIC-AVQA,li2023progressive,yun2021pano,LAVISH_CVPR2023,lao2023coca} employ a typical dual-stream framework. As shown in Fig. \ref{fig2}.a, such architecture processes audio and video in each stream separately, overlooking the natural temporal correlation at the segment level between audio and visual modalities. In particular, the temporal grounding and audio-visual fusion are isolated, with fusion occurring through an additional module. In SOTA \cite{li2023progressive}, they fused audio-visual features on a segment-by-segment scale while overlooking the relative independence of audio and visual content at the video level. The proportion of sensory information relied upon to answer different questions is likely to be different at each timestamp.

To effectively address these challenges and better exploit the pretrained knowledge in VLMs, we propose a CLIP-powered target-aware single-stream (TASS) network for AVQA. Our proposed approach has two key components.

Firstly, we propose a novel target-aware spatial grounding (TSG+) module, which enables the model to focus on visual regions relevant to the query subject (\textit{i.e.}, target), instead of all sound-related visual regions. Specifically, we exploit the explicit semantics of text modality, \textit{i.e.}, the question, and introduce it into audio-to-visual grounding. Besides, we design a new approach to transfer the image-text matching knowledge in large pretrained visual-language models (CLIP \cite{clip}) to our proposed region-level target-aware process. Specifically, we enhance the region-text alignment without any ground-truth labels by adopting an auxiliary loss that leverages the natural matching of audio-visual segments to construct semantic negative samples sampled from different videos. We increase the dissimilarity between text features and regional visual features with different audio attributes, that is, audio serves as a semantic label for images, with different objects possessing different sounds. This provides a new path towards better utilizing the pretrained knowledge in VLMs when only coarse-grained Q\&A labels are available. The CLIP-powered target-aware process facilitates the model to focus on the sounding area of interest, thus improving the overall performance of our question-oriented AVQA model. 

Secondly, we propose a cross-modal synchrony loss (CMS) and a single-stream joint temporal grounding (JTG) module. In contrast to the existing prevalent two-stream frameworks that treat audio and video as separate entities (Fig. \ref{fig2}.a), we design a novel single-stream audio-visual interleaved pattern (Fig. \ref{fig2}.b), which maintains both the integrity of the audio-visual content at the segment level and the relative independence between the audio and visual content at the video level. We leverage \textit{question} as a medium to establish connections between audio and visual features. This extends the image-text matching knowledge in CLIP \cite{clip} to audio-text matching through audio-visual synchronization. Specifically, CMS loss ensures that the attention weights of audio and visual modalities are temporally aligned softly during question-aware temporal grounding, achieved through JS divergence to maintain consistent distributions.  Moreover, it presents avenues to incorporate question-aware temporal grounding and audio-visual fusion into a more straightforward single-stream architecture, instead of the conventional approach of performing temporal grounding first and fusion later. 
% by constraining the consistency of the distribution of text-audio temporal correlations with that of text-video temporal correlations. 
In this way, the network is forced to jointly capture and fuse audio and visual features that are supposed to be united and temporally synchronized. This simpler architecture facilitates competitive performance. 

In addition, we propose a pooling-based feature preprocessing approach before feeding to the model. Video sequences, with significant data and redundancy, contribute to the computational intensity of AVQA, making maximizing accuracy with minimal training resources a main challenge. While most methods \cite{MUSIC-AVQA,lao2023coca,jiang2023avqa} commonly adopt a low fixed sampling rate for video, forming a much shorter frame/audio sequence as the model input, this approach inevitably leads to the loss of video content. The current SOTA method \cite{li2023progressive} proposed a learnable temporal segment selection module to automatically form a shorter sequence of key segments for input to the subsequent network. In contrast to these approaches, our results show that directly pooling the frame/audio feature sequence along the time dimension to obtain a shorter feature sequence as network input not only yields superior performance but also requires comparatively fewer resources.

In summary, we propose a novel end-to-end CLIP-powered TASS-Net for AVQA, achieving new state-of-the-art results. The main contributions can be summarized as follows:

$\bullet$ We propose a novel target-aware spatial grounding (TSG+) module to leverage the pretrained image-text matching knowledge in CLIP models into our region-text matching process by constructing semantic negative audio-visual pairs.

$\bullet$ We propose a single-stream joint temporal grounding (JTG) module, which treats audio and video as a cohesive entity and seamlessly integrates the fusion and temporal grounding. We also propose a cross-modal synchrony loss (CMS) to extend the image-text knowledge to audio-text matching by facilitating the natural temporal synchronization between audio and video. 

$\bullet$ Our proposed video feature preprocessing approach significantly improves the performance with relatively less resource consumption.

% Our proposed approach is evaluated on the MUSCI-AVQA \cite{MUSIC-AVQA} dataset, and experimental results demonstrate its effectiveness and superiority over existing state-of-the-art methods.

This work is built upon our previous conference version \cite{jiang2023avqa}. We substantially revise and significantly extend the previous work in several aspects. First, we newly transfer the implicit prior knowledge from the vision-language pretrained CLIP model to our model without any ground-truth labels by constructing semantic negative audio-visual pairs and leveraging the natural audio-visual matching characteristic. Second, we extend our original TSG module as TSG+ by modifying the target-aware process to obtain a superior CLIP-powered AVQA model. Third, we propose a video feature preprocessing approach to achieve significant performance improvement with as little resource consumption as possible. Fourth, we provide additional technical details, clearer explanations of the contributions/methods and conduct extensive new experiments, including quantitative/qualitative analysis, ablation studies, etc.

\section{Related Works}

\subsection{Vision-Language Models}
Vision-language pre-training has attracted growing attention \cite{lu2019vilbert,lei2021less,coop} during the past few years for working on the interaction of computer vision and natural language processing fields. As a milestone, Radford et al. propose a large-scale vision-language model (VLM) CLIP \cite{clip}, which employs a contrastive learning strategy on a huge amount of image-text pairs and shows impressive transferable ability over 30 classification datasets. Many follow-ups have been proposed to apply it to other finer-grained domains, \textit{e.g.}, DenseCLIP \cite{rao2022denseclip}, StructureCLIP \cite{StructureCLIP}. Usually, additional auxiliary tasks/objectives are required for the features directly extracted by the vanilla CLIP model to better transfer the image-text matching knowledge to the finer-grained downstream tasks. DenseCLIP \cite{rao2022denseclip} converted the original image-text matching problem in CLIP to a pixel-text matching problem and used the pixel-text score maps to guide the learning of dense prediction models. However, there are very few attempts at audio-visual question answering (AVQA) tasks via the CLIP model since only coarse-grained Q\&A labels are available and an additional modality, \textit{i.e.}, audio, is involved. \cite{li2023progressive} applied the CLIP model in the AVQA task for the first time, but they only used CLIP as a feature extractor and did not fully utilize the pre-training knowledge. Differently, we consider the natural audio-visual matching characteristics to better extend the image-text matching knowledge to region-text matching and audio-text matching without using any fine-grained semantic labels.

\subsection{Audio-Visual-Language Learning}
By integrating information from multiple modalities, it is expected to explore a sufficient understanding of the scene and reciprocally nurture the development of specific tasks within a single modality. AVLNet \cite{rouditchenko2020avlnet} and MCN \cite{chen2021multimodal} utilize audio to enhance text-to-video retrieval. AVCA \cite{mercea2022avca} proposes to learn multi-modal representations from audio-visual data and exploit textual label embeddings for transferring knowledge from seen classes of videos to unseen classes. 
% AVSD \cite{schwartz2019avsd,zhu2020describing} takes audio modality into account and explores video comprehension for dialog. 
% VAST \cite{tan2023VAST} utilizes the cross-modal semantic consistency and designs multimodal alignment objectives to achieve audio source separation. 
Compared to previous works in audio-visual learning, such as sounding object localization \cite{afouras2020selfSOL,hu2020discriminativeSOL,hu2022mixSOL}, and audio-visual event localization \cite{liu2022dmin,jiang2023avel,zhou2021psp,xuan2020cross}, these works \cite{mercea2022avca,zhu2020describing,tan2023VAST} have made great progress in integrating the naturally aligned visual and auditory properties of objects and enriching scenes with explicit semantic information by further introducing textual modalities. Besides, there are many works \cite{akbari2021vatt,zellers2022merlot,gabeur2020multi} propose to learn multimodal representations from audio, visual and language modalities that can be directly exploited for multiple downstream tasks. 
% However, these works target learning single or multi-modal representations rather than exploring the basic yet challenging spatio-temporal reasoning ability during scene understanding. In this work, we follow MUSIC-AVQA \cite{MUSIC-AVQA}, which facilitates the study of spatio-temporal reasoning for dynamic and long-term audio-visual scenes, to further exploit the explicit semantic of text modality and the natural alignment of audio-visual cues.
Unlike previous works focused on learning single or multi-modal representations, this work delves into the fundamental yet challenging task of spatio-temporal reasoning in scene understanding. Building upon MUSIC-AVQA \cite{MUSIC-AVQA} benchmark, our approach leverages textual explicit semantics to integrate audio-visual cues to enhance the study of dynamic and long-term audio-visual scenes.

\subsection{Audio-Visual Question Answering}
The demand for multimodal cognitive abilities in AI has grown alongside the advancements in deep learning techniques. Audio-Visual Question Answering (AVQA), unlike previous question answering \cite{lei2018tvqa,you2021self,chen2021learning,wang2021mirtt,vqaijcv}, which exploits the natural multimodal medium of video, is attracting increasing attention from researchers \cite{zhuang2020multichannel,miyanishi2021watch,schwartz2019avsd,zhu2020describing}. 
% Pano-AVQA \cite{yun2021pano} introduces audio-visual question answering in panoramic video and the corresponding Transformer-based encoder-decoder approach. 
AVST \cite{MUSIC-AVQA} offers a strong baseline by decomposing AVQA into audio-visual fusion through spatial correlation of audio-visual elements and question-aware temporal grounding through text-audio cross-attention and text-visual cross-attention. 
% AVQA \cite{yang2022avqa} proposed a hierarchical audio-visual fusing module to explore the impact of different fusion orders between the three modalities on performance. 
LAVISH \cite{LAVISH_CVPR2023} introduced a novel parameter-efficient framework to encode audio-visual scenes, which fuses the audio and visual modalities in the shallow layers of the feature extraction stage. Although LAVISH proposes a robust audio-visual backbone network, it still necessitates the spatio-temporal grounding network proposed in \cite{MUSIC-AVQA}, as MUSCI-AVQA contains dynamic and long-duration scenarios requiring a significant capability of spatio-temporal reasoning. COCA \cite{lao2023coca} proposes a collaborative causal regularization to remedy data biases in the MUSIC-AVQA dataset. PSTP-Net \cite{li2023progressive} proposes to progressively identify key spatio-temporal regions \textit{w.r.t.} questions, which is the first attempt to improve efficiency while retaining video content as much as possible in AVQA. Unlike previous works, we propose a TSG module to leverage the explicit semantic of inquiry target and JTG to leverage the temporal correlation between audio and video in a novel single-stream framework, thus improving the multimodal learning of audio-visual-language.

\section{Methodology}

\begin{figure*}[t]
\centering
\includegraphics[width=16.0cm]{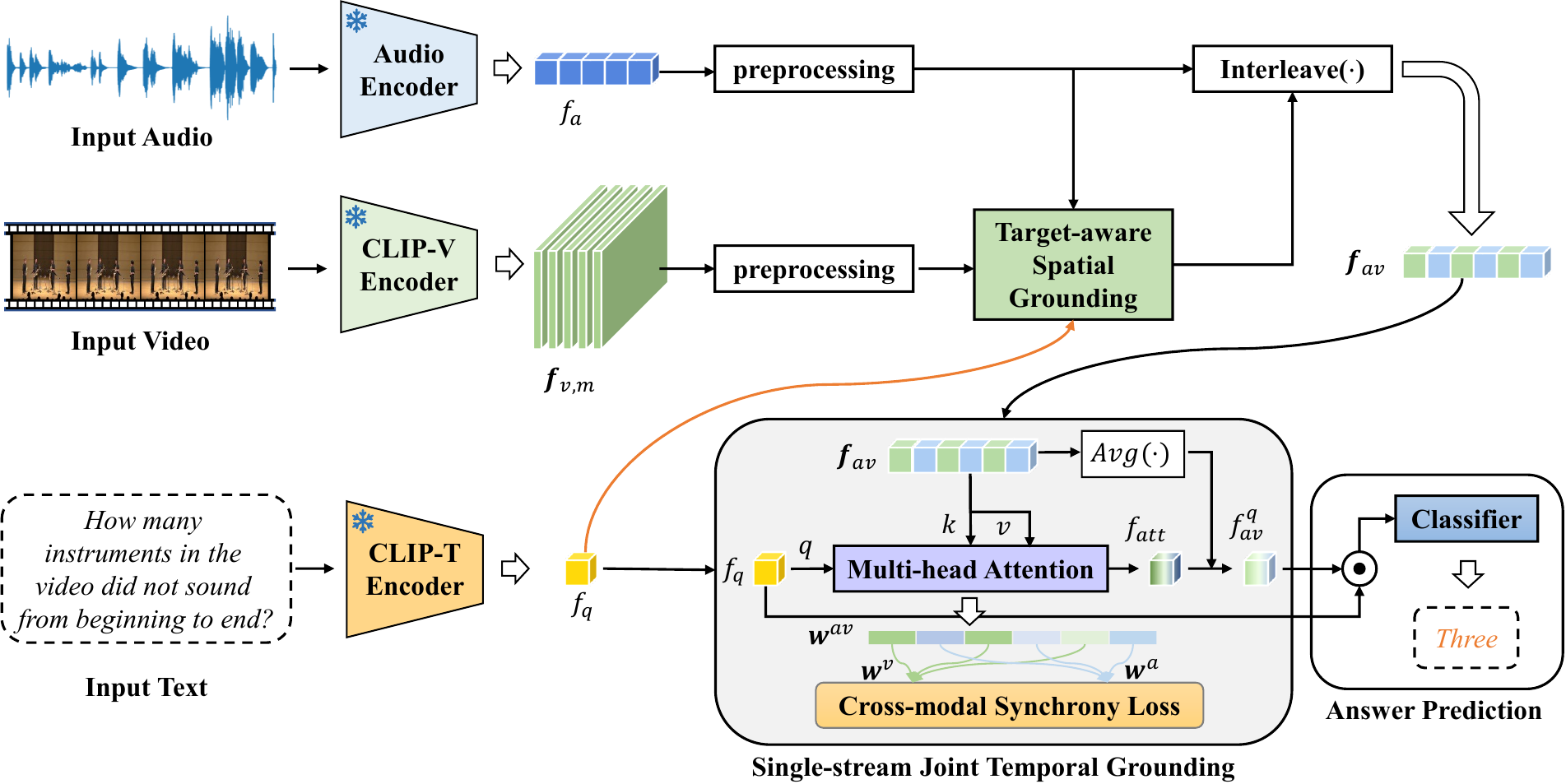}\\
\caption{The proposed target-aware single-stream network. We introduce text modality with explicit semantics into the audio-visual spatial grounding to associate specific sound-related visual features with the subject of interest, i.e., the target. We exploit the proposed cross-modal synchrony loss to incorporate audio-visual fusion and question-aware temporal grounding within a single-stream architecture. Finally, simple fusion is employed to integrate audio-visual and question information for predicting the answer.}\label{fig3} 
\end{figure*}

\subsection{Problem Setup and Model Overview}

\textbf{Problem Setup.} The basic audio-visual question answering problem can be formulated as:

\begin{equation}
\label{eq1}
\begin{aligned}
    a&=\arg\max\limits_{a\in \mathcal{A}}{\text{Pr}(a|A,V,Q)}\\ 
    &= p_\phi(a|\alpha_\theta(\boldsymbol{h}_a,\boldsymbol{h}_v,\boldsymbol{h}_q);\mathcal{A})\\
\end{aligned}
\end{equation}
where $\boldsymbol{h}_a$, $\boldsymbol{h}_v$, $\boldsymbol{h}_q$ is the audio feature, visual feature, and question feature, respectively, which could be extracted from A/V/Q with the pretrained deep neural network models. $\alpha_\theta$ is the multimodal learning model used to integrate features from the input three modalities, and $p_\phi(a|\cdot ;\mathcal{A})$ is the answer predictor probability model to predict the final answer $a$ from the answer set $\mathcal{A}$, with learnable weights $\theta$ and $\phi$, respectively.

\textbf{Model Overview.} To solve the AVQA problem, we propose a CLIP-powered target-aware single-stream network.
% and ensure the integration between audio and visual modalities by observing the natural integrity of the audio-visual cues. 
% The aim is to achieve better audio-visual scene understanding and intentional spatio-temporal reasoning. 
An overview of the proposed framework is illustrated in Fig. \ref{fig3}. Concretely, we start with the off-the-shelf CLIP model for extracting visual and textual features from videos and questions, respectively, while the audio feature is extracted using the corresponding frozen encoder. We apply our proposed parameter-free preprocessing method to process the visual and audio features before feeding them into the network. Within the network, we introduce a target-aware spatial grounding (TSG+) module to transfer the pre-trained image-text matching knowledge from the CLIP model. Next, we interleave visual and audio features to obtain a single-stream feature sequence. Subsequently, we design a single-stream joint temporal grounding (JTG) module to capture and fuse question-relevant audio-visual features simultaneously with the constraints of our proposed cross-modal synchrony (CMS) loss, which also extends the image-text matching knowledge to audio-text matching. Finally, we merge question and audio-visual features, sending them to the classifier for answer prediction.

\subsection{Input Embeddings and Pre-processing}
Given an input video sequence containing both visual and audio tracks, it is first divided into $T_1$ non-overlapping visual and audio segment pairs ${\{V_t, A_t\}}_1^{T_1}$, where each segment is 1-second long. The question sentence \textit{Q} consists of a maximum length of \textit{N} words. For a fair comparison, we use the standard frame sampling rate of 1\textit{fps} to demonstrate the effectiveness of our proposed method.

\textbf{Audio Embedding.} Each audio segment $A_t$ is encoded into $f_a^t\in \mathbb{R}^{d}$ by the pretrained VGGish \cite{gemmeke2017audioset} model, where $d$ is the feature dimension. The VGGish model is a VGG-like 2D CNN network trained on AudioSet \cite{gemmeke2017audio}, employing over transformed audio spectrograms. 
% The audio features can be formulated as $F_a=\{f_a^1,f_a^2,...,f_a^{T_1}\}$.

\textbf{Visual Embedding.} A fixed number of frames are sampled from all video segments. Each sampled frame is encoded into visual feature map $\boldsymbol{f}_{v,m}^t\in \mathbb{R}^{h\times w\times d}$ as in \cite{rao2022denseclip} by the pretrained CLIP \cite{clip} for each segment $V_t$, where $h$ and $w$ are the height and width of the feature maps, respectively. 
% The visual features can be formulated as $F_v=\{f_v^1,f_v^2,...,f_v^{T_1}\}$.

\textbf{Question Embedding.} The question sentence \textit{Q} is tokenized into \textit{N} individual word embeddings ${\{q_n\}}_{n=1}^N$, and then fead into the pretrained CLIP \cite{clip} to obtain a sentence-level question feature $f_q\in \mathbb{R}^{1\times d}$. 

\textbf{Pre-processing.} For every $T_2$ of the $T_1$ audio and visual segments, we utilize global average pooling over temporal dimension to generate one feature vector/map to reduce the amount of data processed by the trainable network while retaining as much video content as possible. Finally, we obtain the audio features as $F_a=[ f_a^1;f_a^2;...;f_a^{T} ]$ and visual features as $\boldsymbol{F}_v=[\boldsymbol{f}_{v,m}^1;\boldsymbol{f}_{v,m}^2;...;\boldsymbol{f}_{v,m}^{T}]$, where $T$ equals $T_1/T_2$.

Noted the used pretrained models are all frozen.

\subsection{Pretrained Knowledge-Guided Target-aware Spatial Grounding}
While sound source localization in visual scenes reflects the spatial association between audio and visual modality, it is cumbersome to elaborately align all audio-visual elements during question answering due to the high complexity of audio-visual scenes. Therefore, we proposed the CLIP-powered target-aware spatial grounding module (TSG+) to encourage the model to focus on the truly interested query object by introducing explicit semantics from the question. Although the absence of precise semantic ground-truth labels hinders effective spatial alignment between text and visual modalities, we design a target-aware process to transfer the pretrained knowledge from CLIP \cite{clip} model to our audio-visual scene understanding.

\begin{figure}[t]
\centering
\includegraphics[width=7.6cm]{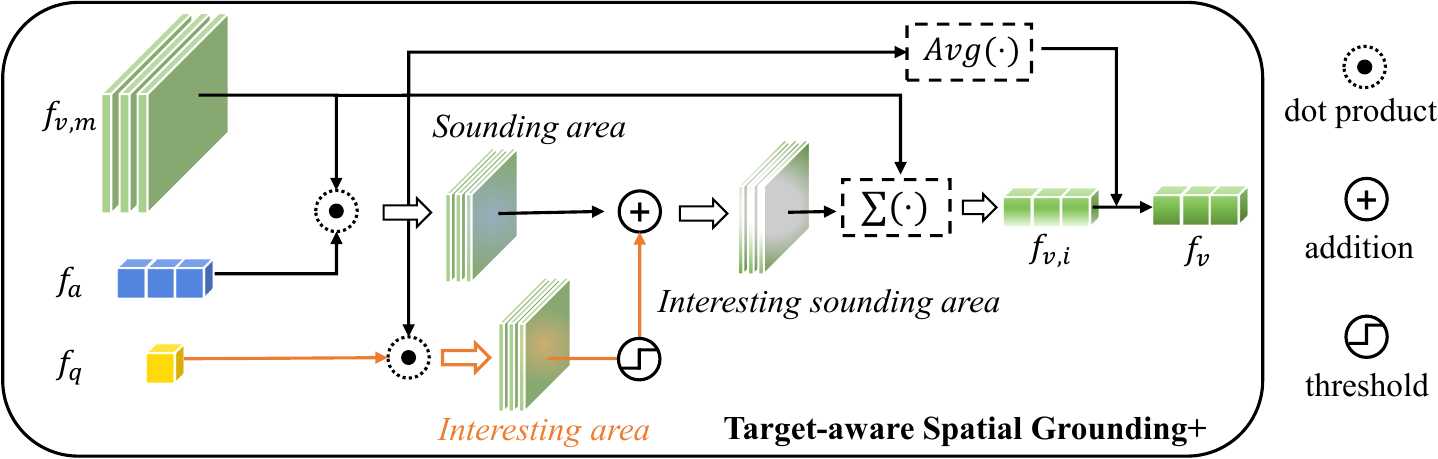}\\
\caption{The illustration of the Target-aware Spatial Grounding module (TSG+), which leverages the explicit semantics from textual modality to conduct visual spatial grounding.}\label{fig3.5} 
\end{figure}

\textbf{Target-aware process.} As shown in Fig. \ref{fig3.5}, for each segment, the visual feature map $\boldsymbol{f}_{v,m}^t$, the audio feature $f_a^t$ and the interesting target feature $f_{tgt}$ compose a matched triplet. Firstly, We maintain the resolution of the $\boldsymbol{f}_{v,m}^t$ to conduct region-level operations and reshape it from $h\times w\times d$ to $hw\times d$. For each triplet, we can compute the interesting sound-related visual features $f_{v,i}^t$ as:
\begin{gather}
    \label{eq5}
f^t_{v,i}=\boldsymbol{f}^t_{v,m}\cdot Softmax(\boldsymbol{s}_a+\boldsymbol{\hat{s}}_q)\\
\boldsymbol{s}_a = Softmax({(f^t_a)}^\top\cdot \boldsymbol{f}_{v,m}^t)\\
\boldsymbol{s}_q = Softmax({(f_{tgt})}^\top\cdot \boldsymbol{f}_{v,m}^t)\\
\boldsymbol{\hat{s}}_q = \boldsymbol{s}_q \mathbb{I}(\boldsymbol{s}_q-\tau)
    % f^t_{v,i}=f^t_{v,m}\cdot\sigma\left[\sigma({(f^t_a)}^\top\cdot f_{v,m}^t)\cdot \sigma({(f_{tgt})}^\top\cdot f_{v,m}^t)\right]
\end{gather}
where $s_a,s_q\in \mathbb{R}^{1\times hw}$, $f^t_{v,i}\in \mathbb{R}^{1\times d}$, $Softmax$ is the softmax function, and $(\cdot)^\top$ represents the transpose operator. In particular, we adopt a simple thresholding operation to better integrate the text modality into an audio-visual scene. Specifically, $\tau$ is the hyper-parameter, selecting the visual areas that are highly relevant to the query subject. $\mathbb{I}(\cdot)$ is an indicator function, which outputs 1 when the input is greater than or equal to 0, and otherwise outputs 0. Next, we add the text-visual attention map to the audio-visual attention map to obtain the target-aware visual attention map. In this way, the TSG+ module will focus on the interesting sounding area instead of all sounding areas. To prevent possible visual information loss, we averagely pool the visual feature map $\boldsymbol{f}_{v,m}^t$, obtaining the global visual feature $f_{v,g}^t \in \mathbb{R}^{1\times d}$. The two visual feature is fused as the visual representation: $f_v^t=\textbf{FC}(\tanh\left[f_{v,g}^t;f_{v,i}^t\right])$, where \textbf{FC} represents fully-connected layers, and $f^t_v\in \mathbb{R}^{1\times d}$.

\textbf{Knowledge Transfer.} Moreover, we adopt an auxiliary objective to better transfer the image-text matching knowledge from CLIP \cite{clip} models to a region-text matching problem in our TSG+ module. Inspired by \cite{MUSIC-AVQA}, we combine the target-aware visual representation $f_v^t$ and audio representation $f_a^t$ to predict whether the audio-visual pairs are matched or not. This separates text features and regional visual features with different audio attributes, that is, audio serves as a semantic label for images, with different objects possessing different sounds. Concretely, we leverage the natural matching of audio and visual to construct positive and negative audio-visual pairs based on whether the audio and target-aware visual representations belong to the same segment. We compute the match loss as:
\begin{gather}
    \label{eq2}
    \mathcal{L}_s=\mathcal{L}_{ce}(y^{m},\hat{y}^t)\\
    \hat{y}=\text{MLP}([f^t_a;f^t_v])
\end{gather}
where $\mathcal{L}_{ce}$ denotes the cross-entropy loss. Concretely, $y^m=1$ when $f_a^t$ and $f_v^t$ are the from matched pair, otherwise $y^m=0$. With this auxiliary objective, we increase the region-level dissimilarity between text features and visual features with different audio attributes, thus better utilizing the image-text matching knowledge from CLIP \cite{clip} model.

\subsection{Pretrained Knowledge-Enhanced Joint Audio-visual Temporal Grounding}

In the natural environment, visual and audio information are different attributes of the same thing, \textit{i.e.}, the two are inseparable. Therefore, we propose a single-stream joint temporal grounding (JTG) module and cross-modal synchrony (CMS) loss to treat the visual modality and audio modality as a whole instead of separate entities as before. This also further utilizes the pretrained knowledge in VLMs trained with image-text matching.

\textbf{Cross-modal synchrony (CMS) loss.} Temporal synchronization is a characteristic of the audio and visual modalities, but in multimedia videos, audio and visual streams do not strictly adhere to simple synchronization. We use the question feature as the intermediary to constrain the temporal distribution consistency of the audio and visual modalities, thus implicitly modeling the synchronization between the audio and video. Concretely, given a video-related question feature $f_q$ and audio-visual features ${\{f_a^t,f_v^t\}}_{t=1}^T$, we first compute the weight of association between the given question and the input sequence, based on how closely each timestamp is related to the question, as:
\begin{gather}
    \label{eq7}
    \boldsymbol{a}_{q}=Softmax(\frac{f_q\boldsymbol{f}_a^\top}{\sqrt{d}})\\
    \boldsymbol{v}_{q}=Softmax(\frac{f_q\boldsymbol{f}_v^\top}{\sqrt{d}})
\end{gather}
where $\boldsymbol{f}_a=\left[f_a^1;\cdots;f_a^T\right]$ and $\boldsymbol{f}_v=\left[f_v^1;\cdots;f_v^T\right]$; $f_q\in\mathbb{R}^{1\times d}$, $\boldsymbol{f}_a\in\mathbb{R}^{T\times d}$, $\boldsymbol{f}_v\in\mathbb{R}^{T\times d}$; $d$ is a scaling factor with the same size as the feature dimension. In this way, we obtain the question-aware weights $\boldsymbol{a}_{q},\boldsymbol{v}_{q}\in\mathbb{R}^{1\times T}$ of audio and video sequence, respectively. 

Next, we employ the Jensen-Shannon (JS) divergence as a constraint. Specifically, the JS divergence measures the similarity between the probability distributions of two sets of temporal vectors, corresponding to the audio and visual question-aware weights, respectively. By minimizing the JS divergence, we aim to encourage the temporal distributions of the two modalities to be as close as possible, thus promoting their question-contributed consistency in the JTG process. The CMS loss can be formulated as:
% \begin{small}
\begin{gather}
    \label{eq10}
    \mathcal{L}_{cms} = \frac{1}{2}D_{KL}(\boldsymbol{a}_{q}\|\boldsymbol{m})+\frac{1}{2}D_{KL}(\boldsymbol{v}_{q}\|\boldsymbol{m})\\
    \boldsymbol{m}=\frac{1}{2}(\boldsymbol{a}_{q}+\boldsymbol{v}_{q})\\
    D_{KL}(\boldsymbol{P}\|\boldsymbol{Q})= {\textstyle \sum_{t}^{T}}P(t)\log{\frac{P(t)}{Q(t)}} 
\end{gather}
% \end{small}
Note that JS divergence is symmetric, i.e., $JS(\boldsymbol{P}||\boldsymbol{Q}) = JS(\boldsymbol{Q}||\boldsymbol{P})$. The weights can be obtained as an additional output of an encapsulated Transformer layer or multi-head attention (MHA), which facilitates easy implementation of CMS loss in different Transformer-based methods.

\textbf{Joint temporal grounding (JTG) module.} Previous approaches to joint audio-visual learning have typically used a dual-stream structure with a decoupled cross-modal fusion module. However, the proposed CMS loss makes single-stream networks for audio-visual learning possible and can naturally integrate the two processes of audio-visual fusion and temporal grounding into one module. Specifically, we first adopt LSTM to model the temporal dependencies in the two modalities respectively. Then, we interleave the video feature tensor and audio feature tensor by segment, which aims to maintain the integrity of the audio-visual content at the segment level, as:
\begin{equation}
    \label{eq13}
    \boldsymbol{f}_{av}=\text{IL}(\boldsymbol{f}_v;\boldsymbol{f}_a)=\left[f_v^1;f_a^1;\cdots;f_v^T;f_a^T\right]
    % \boldsymbol{f}_{av}=\text{IL}(\boldsymbol{f}_a;\boldsymbol{f}_v)=\left[f_a^1;f_v^1;\cdots;f_a^T;f_v^T\right]
\end{equation}
where IL denotes that the features of two modalities are InterLeaved by segments, $\boldsymbol{f}_{av}\in\mathbb{R}^{2T\times d}$ represents the multimedia video features. Next, we perform MHA to aggregate critical question-aware audio-visual features among the dynamic audio-visual scenes as:
% \begin{aligh}
%      \label{eq15}
%     \boldsymbol{f}_{att} &= \sum_{t=1}^{2T}w_t^{av}f_{av}^t \\
%     &=\text{Softmax}((h_q\boldsymbol{W}_q)(\boldsymbol{f}_{av}\boldsymbol{W}_k)^\top)(\boldsymbol{f}_{av}\boldsymbol{W}_v)
% \end{aligh}
% \begin{equation}
%     \label{eq14}
%     \boldsymbol{f}_{att} =\text{MHA}(h_q,\boldsymbol{f}_{av},\boldsymbol{f}_{av})= \sum_{t=1}^{2T}w_t^{av}f_{av}^t
% \end{equation}
% \begin{equation}
%     \label{eq15}
%     \boldsymbol{w}^{av}=\text{Softmax}((h_q\boldsymbol{W}_q)(\boldsymbol{f}_{av}\boldsymbol{W}_k)^\top)
% \end{equation}
\begin{gather}
    \label{eq14}
    f_{att} =\text{MHA}(h_q,\boldsymbol{f}_{av},\boldsymbol{f}_{av})= \sum_{t=1}^{2T}w_t^{av}f_{av}^t\\
     \boldsymbol{w}^{av}=\text{Softmax}((f_q\boldsymbol{W}_q)(\boldsymbol{f}_{av}\boldsymbol{W}_k)^\top)
\end{gather}
\begin{equation}
\label{eq16}
    f_{av}^q = f_{att}+\text{MLP}(Avg(\boldsymbol{f}_{av}))
\end{equation}
where $f_{av}^q\in \mathbb{R}^{1\times d}$ represents the question grounded audio-visual contextual embedding, which is more capable of predicting correct answers. The model will assign higher weights to segments that are more relevant to the asked question in a modality-agnostic manner. Such a modality-agnostic single-stream network structure facilitates the model's emphasis on the segments that have more useful content for answering the question. 

\textbf{Knowledge Extension.} CLIP-encoded text features have greater similarity with visual features compared to audio features. Therefore, we use CMS to extend the knowledge of image-text matching in the CLIP model to audio-text matching by observing audio-visual temporal synchronization. Firstly, to preserve the relative independence between the audio and visual content at the video level, we retrieve the temporal distribution weights for each modality from the multi-head attention (MHA) output. Subsequently, we apply our proposed CMS loss to ensure the audio-visual synchronization at the segment level, thus achieving knowledge extension as follows:
\begin{gather}\label{eq17}
    \mathcal{L}_{cms}=JS(\boldsymbol{w}_a\|\boldsymbol{w}_v)\\
    \boldsymbol{w}_v= {\{\boldsymbol{w}^{av}_{2i}\}}^{2T}_{i=1,\cdots,T}\\
    \boldsymbol{w}_a= {\{\boldsymbol{w}^{av}_{2i-1}\}}^{2T}_{i=1,\cdots,T}
\end{gather}
where $\boldsymbol{w}_a,\boldsymbol{w}_v\in\mathbb{R}^{1\times T}$ are question-aware temporal distribution weights of audio and video, respectively. By leveraging the CMS loss, the proposed JTG module can effectively perform both temporal grounding and audio-visual fusion while considering the soft synchronization between the audio and visual modalities. Besides, this further extends the image-text matching knowledge from CLIP \cite{clip} to our audio-text matching. The resulting single-stream architecture simplifies the overall system and treats audio and video as a whole.

\subsection{Answer Prediction}
To verify the audio-visual fusion of our proposed single-stream joint temporal grounding module, we employ a simple element-wise multiplication operation to integrate the question features $f_q$ and the previously obtained audio-visual features $f_{av}^q$. Concretely, it can be formulated as:
\begin{equation}
    \label{eq20}
    e = f_{av}^q \odot f_q
\end{equation}
Next, we aim to choose one correct answer from a pre-defined answer vocabulary. We utilize a linear layer and softmax function to output probabilities $p\in\mathbb{R}^C$ for candidate answers. With the predicted probability vector and the corresponding ground-truth label 
$y$, we optimize it using a cross-entropy loss: $\mathcal{L}_{qa} = - {\textstyle \sum_{c=1}^{C}}y_c\log(p_c) $. During testing, the predicted answer would be $\hat{c} = \arg \max\limits_c(p)$.

\begin{table*}[t]
\caption{Comparisons with state-of-the-art methods on the MUSIC-AVQA dataset. The top-2 results are highlighted. * indicates the length of the video feature sequence input to the network.}\label{tab1}
\centering
\resizebox{\textwidth}{!}{
\begin{tabular}{clcccccccccccccc}
\toprule
\multirow{2}{*}{Encoder} & \multirow{2}{*}{Method} & \multicolumn{3}{c}{Audio Question} & \multicolumn{3}{c}{Visual Question} & \multicolumn{6}{c}{Audio-Visual Question}                          & All   \\
& & Count.   & Comp.  & Avg.   & Count.    & Loc.    & Avg.    & Exist. & Loc. & Count. & Comp. & Temp. & Avg.  & Avg.  \\ \midrule
\multirow{9}{*}{\begin{tabular}[c]{@{}c@{}}A: VGGish\\ V: ResNte-18\\ Q: GloVe \\ (\textit{T}=10)*\end{tabular}} 
& CONVLSTM \cite{fayek2020aqa}   & 74.07    & 68.89    & 72.15  & 67.42   & 54.56   & 60.94   & 82.91          & 50.81  &  63.03         & 60.27   & 51.58  & 62.24 & 63.65 \\
& MCAN \cite{yu2019vqa}          & 77.50    & 55.24    & 69.25  & 71.56   & 70.93   & 71.24   & 80.40          & 54.48  &  64.91         & 57.22   & 47.57             & 61.58 & 65.49 \\
& HME \cite{fan2019videoqa}      & 74.76    & 65.36    & 70.61  & 67.97   & 69.46   & 68.76   & 80.30          & 53.18  &  63.19        & 62.69   & 59.83             & 64.05 & 66.45 \\
& AVSD \cite{schwartz2019avsd}   & 72.41    & 62.46    & 68.78  & 66.00   & 74.53   & 70.31   & 80.77          & 57.93  &  64.03         & 62.85   & 61.07             & 65.44 & 67.32\\
& Pano-AVQA \cite{yun2021pano}   & 75.71    & 65.99    & 72.13  & 70.51   & 75.76   & 73.16   & 82.09          & 61.30  &  65.38         & 63.67   & 62.04             & 66.97 & 69.53 \\
& AVST \cite{MUSIC-AVQA}         & 77.78    & 67.17    & 73.87  & 73.52   & 75.27   & 74.40   & 82.49          & 64.24  &  69.88         & 64.67   & \textbf{65.82}    & 69.53 & 71.59 \\
& COCA \cite{lao2023coca}        & 79.94    & 67.68    & 75.42  & 75.10   & 75.43   & 75.23   &\textbf{83.50}  & \textbf{66.63}  & 69.72 & 64.12   & \underline{65.57} & 69.96 & 72.33 \\
& TJSTG \cite{jiang2023avqa} (Ours)  & \underline{80.38}     & \underline{69.87}    & \underline{76.47}       & \textbf{76.19}  & \underline{77.55}            & \underline{76.88}        & \underline{82.59}            & 64.24         &  \underline{71.54}        & \textbf{66.21}            & 64.84         & \underline{70.13}      & \underline{73.04}      \\
& TASS (Ours)            & \textbf{81.71}            & \textbf{70.37}              & \textbf{77.53}       & \underline{76.02}            & \textbf{78.37}            & \textbf{77.21}        & 82.09            & \underline{66.09}         &  \textbf{72.73}        & \underline{65.94}            & 65.45         & \textbf{70.70}      & \textbf{73.63} \\ \midrule
\multirow{2}{*}{\begin{tabular}[c]{@{}c@{}}A: VGGish \\V/Q: CLIP\end{tabular}} 
& PSTP \cite{li2023progressive} (\textit{T}=60)*   & \underline{73.97}      & \textbf{65.59}        & \underline{70.91}  & \underline{77.15}       & \underline{77.36}       & \underline{77.26}   & \underline{76.18}       & \textbf{71.80}    &  \underline{73.23}   & \textbf{71.79}       & \underline{69.00}    & \textbf{72.57} & \underline{73.52} \\
& TASS (Ours) (\textit{T}=10)*           & \textbf{83.38}            & 63.13              & \textbf{75.92}       & \textbf{80.37}            & \textbf{79.51}            & \textbf{79.93}        & \textbf{82.39}            & \underline{68.91}         & \textbf{75.89}        & \underline{64.40}            & \textbf{69.22}         & \underline{72.33}      & \textbf{74.98} \\
\bottomrule
\end{tabular}}
\end{table*}

\section{Experiments}

Following \cite{MUSIC-AVQA,lao2023coca,jiang2023avqa,li2023progressive}, we evaluate our method on the MUSIC-AVQA dataset. This section presents the setup details and extensive experimental results. We also discuss the model’s performance and specify the effectiveness of each sub-module through ablation studies and qualitative results.

\subsection{Experiments Setting}
\textbf{Dataset.} MUSIC-AVQA dataset \cite{MUSIC-AVQA} contains 45,867 question-answer pairs distributed in 9,288 videos for over 150 hours. Each video contains around 5 QA pairs on average. The questions cover 9 types including three different scenarios, i.e., audio, visual, and audio-visual. The MUSIC-AVQA dataset is well-suited for studying spatio-temporal reasoning for dynamic and long-term audio-visual scenes. 

\textbf{Metric.} Answer prediction accuracy. We evaluate the model's performance in answering different types of questions.

\textbf{Implementation details.} 
The sampling rates of sounds and frames are 16 \textit{kHz} and 1 \textit{fps}. We divide the video into non-overlapping segments of the same length with 1\textit{s}-long. For each video segment, we use the pretrained CLIP-RN50 \cite{clip} to generate the visual feature of size 7×7×2048 for each frame, which is then projected to 7×7×512-D with a linear layer. For each audio segment, we project the extracted 128-D VGGish \cite{gemmeke2017audio} feature into a 512-D feature vector. For each question sentence, we extract its feature by the pre-trained CLIP-RN50 \cite{clip} and obtain a 512-D feature vector. For preprocessing, we set the input length \textit{T} of the video feature sequence to 10 following previous settings \cite{MUSIC-AVQA,jiang2023avqa}. Batch size and number of epochs are 64 and 30, respectively. The initial learning rate is 2e-4 and will drop by multiplying 0.1 every 12 epochs. Our network is trained with the Adam optimizer. We use the \textit{torchsummary} library to calculate the model's parameters. Our model is trained on an NVIDIA GeForce GTX 1080 and implemented in PyTorch.

\textbf{Training Strategy.} Previous methods \cite{MUSIC-AVQA} use a two-stage training strategy, training the targetless spatial grounding module first with the audio-visual pair matching task. We utilize their pretrained weights directly for initializing certain layers that overlap with our approach to present an end-to-end framework. 
% Note that the visual encoder used in \cite{MUSIC-AVQA} is ResNet18 \cite{he2016deepres} which is different from our CLIP-V (ResNet50) encoder, but. 
Finally, We propose to use $\mathcal{L} = \mathcal{L}_{qa}+\mathcal{L}_{cms} + \lambda\mathcal{L}_s$ for AVQA training in an end-to-end manner, where $\lambda$ is 0.5 following previous setting \cite{MUSIC-AVQA}.

\begin{table*}[htb]
\caption{ Ablation studies of different modules on MUSIC-AVQA dataset. The top-2 results are highlighted.}\label{tab2}
\centering
\resizebox{\textwidth}{!}{
\begin{tabular}{cccccccccccccc}
\toprule
\multirow{2}{*}{Method} & \multicolumn{3}{c}{Audio Question} & \multicolumn{3}{c}{Visual Question} & \multicolumn{6}{c}{Audio-Visual Question}                          & All   \\
                        & Count.   & Comp.  & Avg.   & Count.    & Loc.    & Avg.    & Exist. & Loc. & Count. & Comp. & Temp. & Avg.  & Avg.  \\ \midrule
w/o T-A                    & 82.89      & 63.30        & 75.67  & 79.45       & 78.69       & 79.07  & \underline{83.20}       & 66.85    & 75.34    & 62.31       & \textbf{69.34}    & \underline{71.55} & \underline{74.27} \\
w/o $\mathcal{L}_{s}$  & \underline{83.78}    & \underline{63.47}   & \underline{76.29}   & 78.28     & 74.61    & 76.42   & \textbf{83.30}        & 63.26     & 73.99     & 59.67        & 64.60   &69.25  & 72.40 \\
w/o $\mathcal{L}_{cms}$    & 83.28      & 62.79        & 75.73  & \textbf{80.70}       & \underline{79.02}  & \underline{79.85}      & 81.98   & 66.30   & \textbf{76.13}    & 62.03   & 67.64       & 71.08     & 74.22 \\
w/o TSG+                    & 83.68      & \textbf{64.31}        & \textbf{76.54}  & 79.62       & 77.47       & 78.53  & 81.07       & 65.11    & \underline{76.21}    & 62.22       & 67.40    & 70.70 & 73.81 \\
w/o JTG    & \textbf{83.78}      & 62.29        & 75.85  & 78.61       & 78.61       & 78.61   & 82.19       & \underline{67.28}    & 73.68    & \underline{62.67}       & 66.18    & 70.58 & 73.64 \\
TASS(Ours)       & 83.38            & 63.13              & 75.92       & \underline{80.37}            & \textbf{79.51}            & \textbf{79.93}        & 82.39            & \underline{68.91}         & 75.89        & \textbf{64.40}            & \underline{69.22}         & \textbf{72.33}      & \textbf{74.98}\\ \bottomrule
\end{tabular}}
\end{table*}

\subsection{Comparisons with SOTA Methods}
We challenge our method TASS against previous SOTA methods on AVQA. For a fair comparison, We conduct experiments under both common feature encoder settings. As shown in Table \ref{tab1}, We compare our TASS approach with the AQA method \cite{fayek2020aqa}, VQA method \cite{yu2019vqa}, VideoQA method \cite{fan2019videoqa} and current AVQA methods \cite{schwartz2019avsd,yun2021pano,MUSIC-AVQA,lao2023coca,jiang2023avqa,li2023progressive} methods. 

Our CLIP-powered TASS method achieved the highest accuracy of 74.98\%, significantly outperforming the second-best method PSTP \cite{li2023progressive} by 1.46\%$\uparrow$. We achieve a competitive accuracy of 72.33\% on audio-visual questions with a more straightforward single-stream architecture and surpass PSTP by an average of 5.91\%$\uparrow$ and 2.73\%$\uparrow$ on audio and visual questions respectively. In particular, our method shows clear superiority when answering counting questions (average of 5.10\%$\uparrow$), which requires a high conceptual understanding and reasoning ability over object-level information. The improvement over counting questions achieved by our methods (TJSTG \cite{jiang2023avqa} and TASS) can also be observed under the settings of backbone networks ResNet-18 \cite{he2016deepres} (pretrained on ImageNet \cite{russakovsky2015imagenet}) and GloVe \cite{mikolov2013word2vec}. This considerable improvement can be attributed to our proposed TSG+ module, which introduces textual modalities with explicit semantics into the audio-visual spatial grounding process. 

Compared to the second-best TJSTG method on ResNet-18 and GloVe features, \textit{i.e.}, our previous conference version \cite{jiang2023avqa}, our TASS network improved the target-aware process (from TSG to TSG+, see section \ref{sec:a} for details) and added feature preprocessing, improving the accuracy by 0.58\% (from 73.05\% to 73.63\%). The results achieved by our TASS method on ResNet-18 and GloVe features have also exceeded the PTSP method on CLIP features by 0.11\% (from 73.52\% to 73.63\%). It should be noted that the second-best CLIP-based method PSTP \cite{li2023progressive} inputs the video (audio and visual) feature sequence of length \textit{60} to the trainable network while we input the video (audio and visual) feature sequence of length \textit{10} as others. This demonstrates the effectiveness of our preprocessing approach, which achieves better performance without using more training resources (further confirmed by Table \ref{tab3}). In summary, the TASS-Net offers significant improvements over existing approaches and provides a novel insight for question-oriented audio-visual scene understanding.

\subsection{Ablation studies} 
\label{sec:a}
\textbf{The effectiveness of the different modules in our model.} To verify the effectiveness of the proposed components, we remove them from the primary model and re-evaluate the new model. Table \ref{tab2} shows that after removing a single component, the overall model’s performance decreases, and different modules have different performance effects. Firstly, when we remove the target-aware process (denoted as ``w/o T-A'') and use traditional audio-only visual spatial grounding, the accuracy decreases by 0.15\%, 0.86\%, 0.78\%, and 0.71\% under audio, visual, audio-visual, and all questions, respectively. This shows that it is essential to have a targeting process before feature aggregation instead of attending to all the audio-visual cues. Secondly, we remove the match loss (denoted as ``w/o $\mathcal{L}_{s}$''), and the overall accuracy drops to 72.40\%, exhibiting a 2.58\% decline from our full model. ``w/o $\mathcal{L}_{s}$'' is 1.41\% lower the accuracy of removing the entire TSG+ module along with its corresponding $\mathcal{L}_{s}$ (denoted as ``w/o TSG+''). This shows that directly introducing CLIP into the AVQA task without any constraints will not only fail to make full use of its pre-trained text-image matching knowledge, but will also affect the overall performance of the model because the audio features are isolated by the matched image text. Next, we remove the proposed cross-modal synchrony loss (denoted as ``w/o $\mathcal{L}_{cms}$''), and the overall accuracy drops to 74.22\% (0.76\% below our full model). This demonstrates the importance of maintaining the temporal correlation of audio and visual features during the feature selection and fusion process. Moreover, we removed the entire TSG+ module (with its $\mathcal{L}_{s}$) and JTG module (with its $\mathcal{L}_{cms}$) from the full model respectively, both resulting in significant performance degradation. These results show that every component plays an essential role in our TASS network for AVQA.

\begin{table}
\caption{Effect of the preprocessing approach.}
\centering
\setlength\tabcolsep{0.8mm}{
\begin{tabular}{llcccc}
\toprule
Encoder     & Method         & A Avg. & V Avg. & AV Avg. & All   \\
\midrule
\multirow{2}{*}{\begin{tabular}[c]{@{}l@{}}A:VGGish\\ V:ResNet-18\end{tabular}} 
& w/o prep. & 76.41  & 77.04  & 70.09   & 73.05 \\
& w/ prep. & \textbf{77.53}  & 77.21  & 70.70   & 73.63 \\
\midrule
\multirow{2}{*}{\begin{tabular}[c]{@{}l@{}}A:VGGish\\ V:CLIP-V\end{tabular}}    
& w/o prep. & 75.48  & 79.40  & 71.13   & 74.09 \\
& w/ prep.  & 75.92  & \textbf{79.93}  & \textbf{72.33}   & \textbf{74.98}\\
\bottomrule
\end{tabular}
}
\label{tab3}
\end{table}

\textbf{Effect of the preprocessing approach.} As shown in Table \ref{tab3}, the experimental results prove that our preprocessing method has achieved performance improvements under different feature encoder settings. Specifically, it improves accuracy by 0.58\% (from 73.05\% to 73.63\%) and 0.89\% (from 74.09\% to 74.98\%), respectively. Note that it did not cause any increase in training parameters by directly pooling the extracted frame/audio feature sequence along the time dimension to obtain a shorter feature sequence as network input. Compared to the common setting \cite{MUSIC-AVQA,lao2023coca,jiang2023avqa} of sampling video at a low sampling rate to form a shorter visual/audio sequence as the model input, our approach retains more video content with the same resources during training. Compared to the current SOTA method PSTP \cite{li2023progressive} that designed a trainable temporal segment selection module to automatically form a shorter sequence of segments with key content related to the question, our approach achieves superior performance without using more training resources, surpassing it by 1.46\%, as shown in the Table \ref{tab1}. We believe that this preprocessing method can be widely applied to computationally intensive video understanding tasks. It has not only been proven effective in discriminative tasks such as audio-visual event localization \cite{tian2018avel,jiang2023avel}, but is also suitable for fine-grained spatio-temporal reasoning tasks such as audio-visual question answering.

\begin{table}
\caption{Effect of the Target-aware process.}
\label{tab4}
\centering
\setlength\tabcolsep{1mm}{
\begin{tabular}{lccccc}
\toprule
Method & A Avg. & V Avg. & AV Avg. & All   \\
\midrule
TSG w/ \textit{mul}   & 75.85  & 79.77   & 71.80 & 74.63  \\
TSG w/ \textit{max}      & \textbf{76.54} & 79.81 & 71.45 & 74.56 \\
TSG+ w/ \textit{add} (Ours)  & 75.92  & \textbf{79.93}  & \textbf{72.33}   & \textbf{74.98} \\
\bottomrule
\end{tabular}
}
\end{table}

\textbf{Effect of target-aware process.} 
As shown in Table \ref{tab4}, we adopt different ways to introduce question information during the target-aware process, thus verifying the effectiveness of our proposed target-aware spatial grounding (TSG+) module. Specifically, we perform Hadamard product (denoted as ``TSG w/ \textit{mul}'') on the audio-visual attention map $s_a$ and question-visual attention map $s_q$ to conduct the target aware process upon all sounding regions as in our previous version of TSG \cite{jiang2023avqa}. Besides, the threshold value $\tau$ in Hadamard product settings is set to a relatively low 0.02 based on our experience in our previous work. We also adopt maximum selection (denoted as ``TSG w/ \textit{max}'') between $s_a$ and $s_q$, trying to obtain visual features that are significantly related to both audio and text.
% We also adopt two learnable parameters to conduct weighted summation on $s_a$ and $s_q$ (denoted as ``TSG w/\textit{wsum}''), \textit{i.e.}, $f^t_{v,i}=\boldsymbol{f}^t_{v,m}\cdot Softmax(w_1\cdot\boldsymbol{s}_a+w_2\cdot\boldsymbol{\hat{s}}_q)$. 
Compared to these approaches, our approach (denoted as ``TSG w/ add'') achieves the highest accuracy of 74.98\%. All three different target-aware settings achieve higher accuracy than removing the target-aware process (denoted as ``w/o T-A'' in Table \ref{tab2}). The experimental results prove the superiority of our proposed TSG+ module and further demonstrate our target-aware process's effectiveness during the audio-visual learning stage.

\begin{table}
\caption{Impact of various values of $\tau$.}
\label{tab-tau}
\centering
\setlength\tabcolsep{2.5mm}{
\begin{tabular}{lccccc}
\toprule
Method & A Avg. & V Avg. & AV Avg. & All   \\
\midrule
$\tau=0.000$            & 76.29   & 79.40   & 70.96  & 74.14   \\
$\tau=0.005$            & 76.29   & 79.40   & 70.96  & 74.14   \\
$\tau=0.020$            & \textbf{76.41}   & \textbf{80.22}   & 71.04 & 74.42 \\
$\mathbf{\tau=0.025}$       & 75.92  & 79.93  & \textbf{72.33}   & \textbf{74.98} \\
$\tau=0.030$       & 76.04   &80.06   & 71.02 & 74.30 \\
\bottomrule
\end{tabular}
}
\end{table}
In addition, we explore the impact of hyperparameter $\tau$ on model performance. As Table \ref{tab-tau} shows, $\tau$ plays a role in selecting visual areas that are highly relevant to the query subject. The highest accuracy of 74.98\% is achieved when $\tau=0.025$. Note that the range of threshold selection depends on the mean of the attention map, calculated by dividing 1 by the size of the feature map, \textit{i.e.}, $1/(h\times w)$, where $h=w=7$. The results verify the robustness of our target-aware process. The performance decreased by 0.58\% (from 74.98\% to 74.30\%) when the selecting threshold was too high ($\tau=0.030$), which proves the importance of retaining the target information. In addition, $\tau=0.000$ and $\tau=0.005$ have the same performance of 74.14\% (performance decreased by 0.88\%), because when $\tau$ is too small, the threshold operation loses its effect. This demonstrates that in the absence of corresponding semantic labels, the threshold operations over cross-modal alignment can improve the overall performance of the system to a certain extent.
\begin{table}
\caption{Effect of the single-stream pattern on the accuracy(\%). ``IL'' denotes that audio and visual features are interleaved by segments. ``Cat'' denotes that audio and visual features are concatenated by modalities.}
\label{tab5}
\centering
\setlength\tabcolsep{1.8mm}{
\begin{tabular}{lccccc}
\toprule
Method & A Avg. & V Avg. & AV Avg. & All   \\
\midrule
Cat(A;V)  & \textbf{76.35}   & 79.19   & 70.66  & 73.93  \\
Cat(V;A)  & 75.61   & 79.73   & 71.45  & 74.38 \\
IL(A;V)   & 75.67   & 79.69   & 71.41  & 74.36 \\
IL(V;A)  (Ours)  & 75.92  & \textbf{79.93}  & \textbf{72.33}   & \textbf{74.98} \\
\bottomrule
\end{tabular}
}
\end{table}
\begin{table}[t]
\caption{Comparison of dual-stream structure and singe-stream structure.}
\label{tab6}
\centering
\setlength\tabcolsep{2.8mm}{
\begin{tabular}{lccc}
\toprule
\thead[l]{Method} & \thead{$\mathcal{L}_{cms}$} & \thead{Trainable \\ Param.(M)$\downarrow$} & \thead{Accuracy \\ (\%)$\uparrow$} \\
\midrule
\multirow{2}{*}{\begin{tabular}[c]{@{}l@{}}Dual-stream\\ \cite{MUSIC-AVQA}\end{tabular}}   & $\times$    &    12.5            &   72.25       \\
                                   &  $\checkmark$   &   12.5         &   73.68      \\
\midrule
\multirow{2}{*}{\begin{tabular}[c]{@{}l@{}}Single-stream\\ (Ours)\end{tabular}} & $\times$  &   10.9        &  74.22        \\
                                   &  $\checkmark$   & \textbf{10.9}  &  \textbf{74.98}   \\
\bottomrule
\end{tabular}
}
\end{table}

\textbf{Effect of single-stream structure.} We validate the effectiveness of our designed specialized audio-visual interleaved pattern, \textit{i.e.}, IL(V;A), which maintains both the integrity of the audio-visual content at the segment level and the relative independence between the audio and visual content at the video level. As shown in Table \ref{tab5}, we explore different ways of arranging visual and audio features, and the accuracy of our interleaved-by-segments pattern (denoted as ``$\text{IL}(\cdot)$'') is 0.415\% higher on average than the concatenated-by-modals pattern (denoted as ``$\text{Cat}(\cdot)$''). we also conduct a comprehensive comparison between single-stream and dual-stream networks. During the temporal grounding, we switch our single-stream structure to the prevalent two-stream network in \cite{MUSIC-AVQA,lao2023coca}, but still with our proposed TSG+ module and cross-modal synchrony loss, which is denoted as ``Dual-stream'' in Table \ref{tab6}. As shown in Table \ref{tab6}, the ``Single-stream'' that omits the additional fusion module yields 0.85\% higher accuracy with 1.6M fewer parameters than the ``Dual-stream''. This indicates the superiority of single-stream networks over two-stream networks, which utilize the integration of the audio and visual modalities to simultaneously accomplish question-aware temporal grounding and audio-visual fusion.

\textbf{Effect of cross-modal synchrony loss.} 
As shown in Table \ref{tab5} and \ref{tab6}, we further consider the effect of cross-modal synchrony loss under different combination settings of multimedia video features $\boldsymbol{f}_{av}$. In our previous work \cite{jiang2023avqa}, the experiment results on ResNet-18 and GloVe features showed that although the visual-audio order (V;A) always achieves better results than the audio-visual order (A;V), there was no significant gap within the constraints of the CMS. However, as shown in Table \ref{tab5}, our TASS method on CLIP features significantly increased the gap between the different order in which the audio-visual features are combined. Specifically, in different single patterns (concatenated and interleaved), visual-audio sequences (V;A) are on average 0.535\% higher in accuracy than audio-visual sequences (A;V). We believe that the closer similarity between CLIP-encoded question features and visual features is the main reason for this bias. Besides, as depicted in \ref{tab6}, the CMS not only enhanced the accuracy by 0.76\% for our single-stream structure but also demonstrated a significant improvement of 1.43\% for the dual-stream structure network like \cite{MUSIC-AVQA,lao2023coca} (from 72.25\% to 73.68\%). These results validate the robustness and effectiveness of our proposed CMS and the importance of maintaining temporal correlations of audio and visual content in audio-visual learning.

\begin{figure}[t]
\centering
\includegraphics[width=7.4cm]{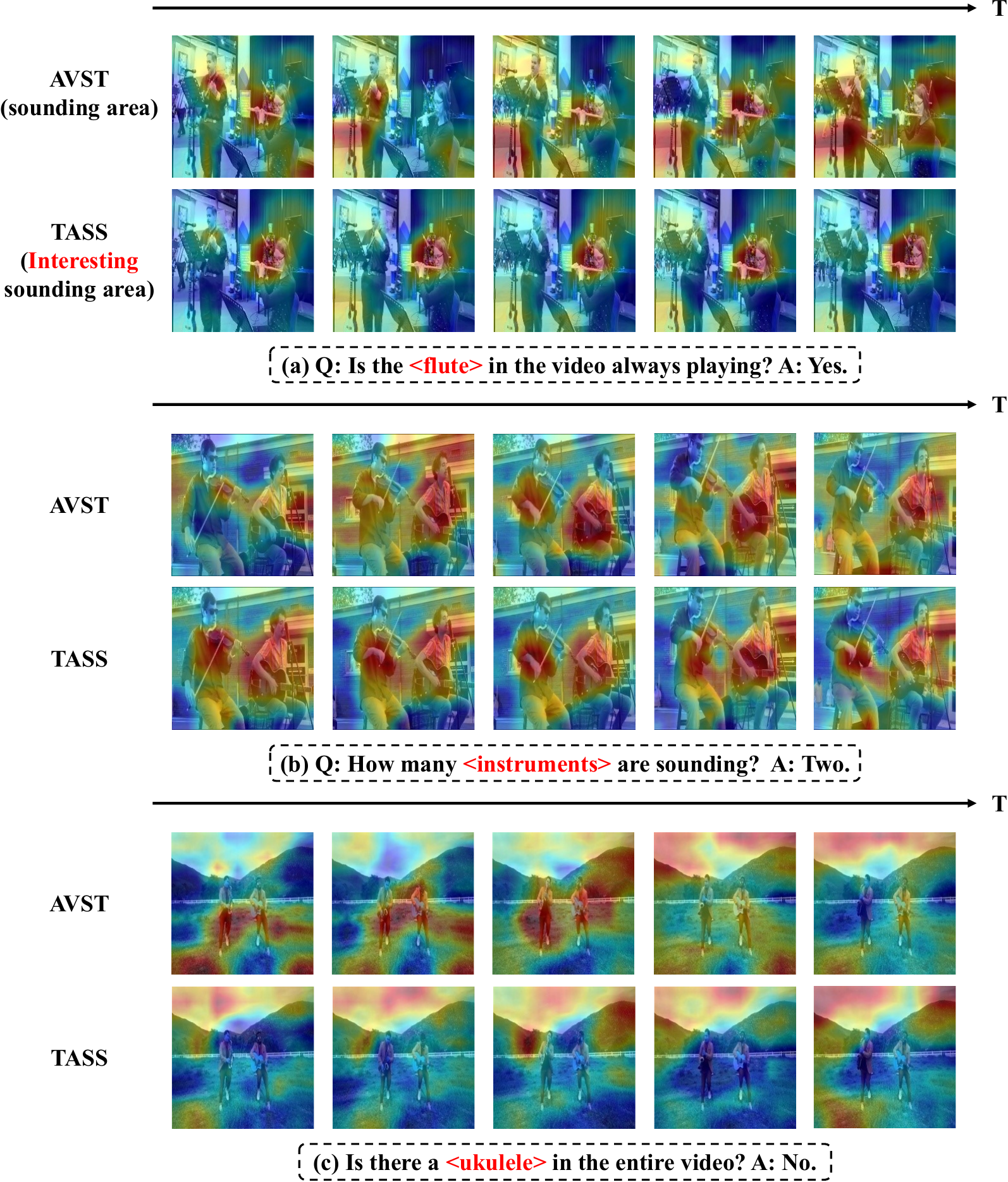}\\
\caption{Visualized target-aware spatial grounding results. Based on the grounding results of our method, the sounding area of interest are accordingly highlighted in spatial perspectives in different cases (a-c), respectively, which indicates that our method can focus on the query subject, facilitating the question-oriented scene understanding and reasoning.}\label{fig4} 
\end{figure}

\subsection{Qualitative analysis} 

\textbf{Target-aware Process.} In Fig. \ref{fig4}, we provide several visualized target-aware spatial grounding results. The heatmap indicates the location of the interesting-sounding source. Note that for a fair comparison, our results were obtained under the same ResNet-18 backbone as AVST \cite{MUSIC-AVQA} which also has a spatial grounding module. Through the results, the sounding targets are visually captured, which can facilitate spatial reasoning. For example, in the case of Fig. \ref{fig4}.(a), compared to AVST \cite{MUSIC-AVQA}, our proposed TASS method can focus on the target, i.e., the \textit{flute}, during spatial grounding. The TSG module offers information about the interesting-sounding object in each timestamp. In the case of Fig. \ref{fig4}.(b) with multiple sound sources related to the target, \textit{i.e.}, \textit{instruments}, our method also indicates a more accurate spatial grounding compared to AVST \cite{MUSIC-AVQA}. When there is no target of interest in the video, as shown in Fig. \ref{fig4}.(c), i.e., the \textit{ukulele}, it can be seen that our method presents an irregular distribution of spatial grounding in the background region instead of the undistinguished sounding area of the \textit{guitar} and \textit{bass} presented by AVST \cite{MUSIC-AVQA}. Furthermore, the JTG module aggregates the information of all timestamps based on the question. These results demonstrate that our proposed method can focus on the most question-relevant audio and visual elements, leading to more accurate question answers.

\begin{figure*}[t]
\begin{minipage}[b]{0.5\linewidth}
  \centering
  \centerline{\includegraphics[width=7.9cm]{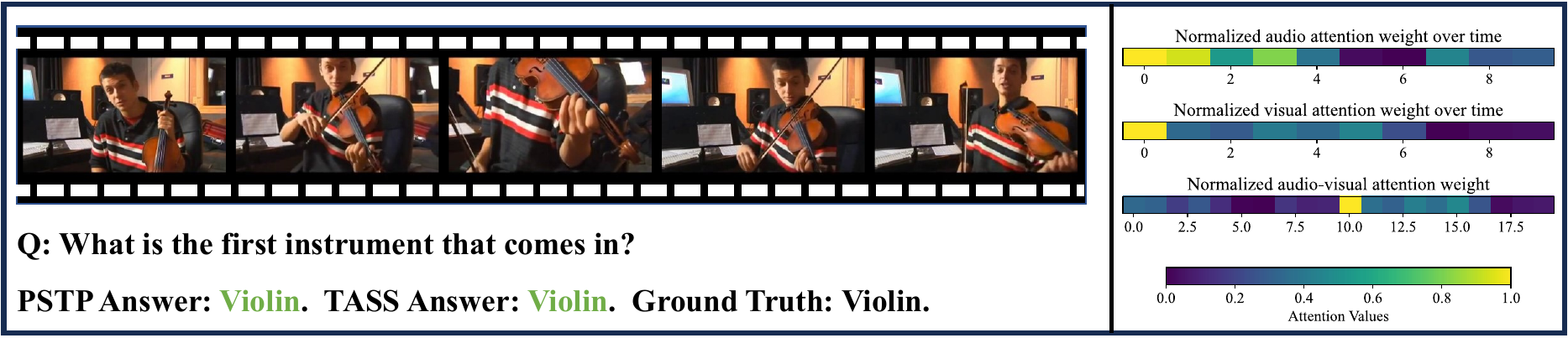}}
  (a) AV-Temporal.
  \label{av_t}
  \vspace{0.2cm}
\end{minipage}
\begin{minipage}[b]{0.5\linewidth}
  \centering
  \centerline{\includegraphics[width=7.9cm]{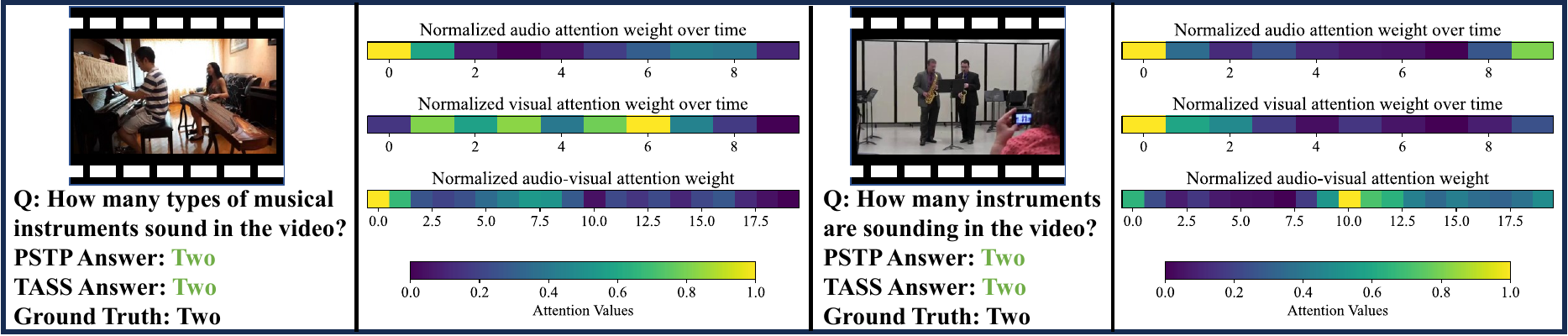}}
  (b) AV-Count.
  \label{av_c}
  \vspace{0.2cm}
\end{minipage}
\begin{minipage}[b]{0.5\linewidth}
  \centering
  \centerline{\includegraphics[width=7.9cm]{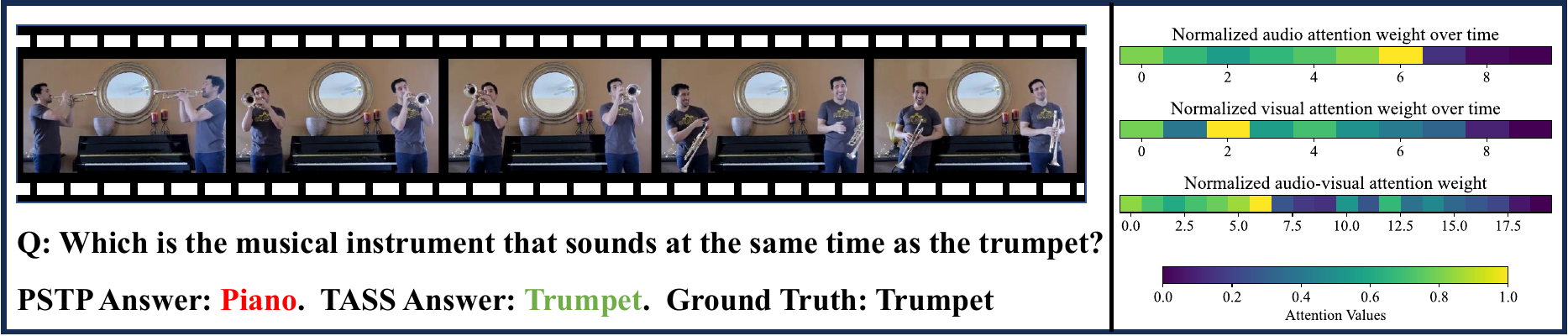}}
  (c) AV-Location.
  \label{av_l}
  \vspace{0.2cm}
\end{minipage}
\begin{minipage}[b]{0.5\linewidth}
  \centering
  \centerline{\includegraphics[width=7.9cm]{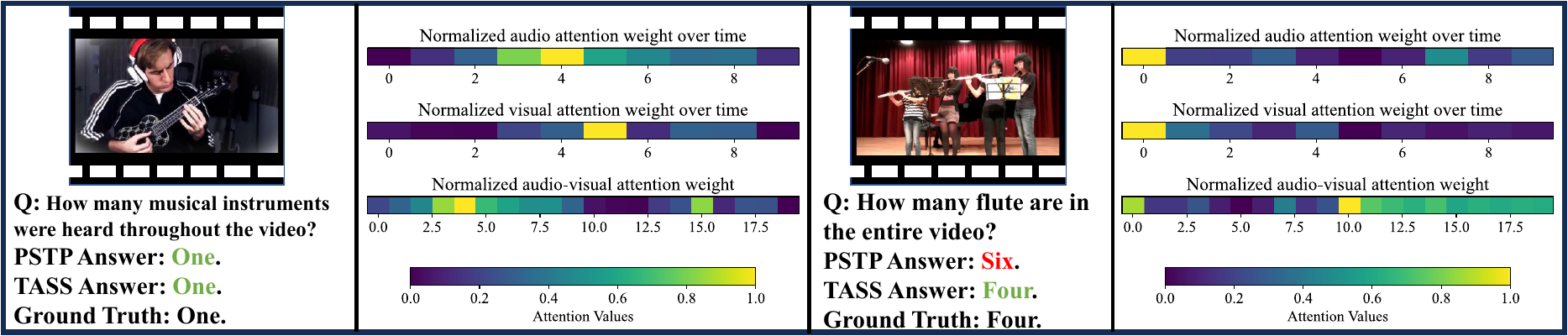}}
  (d) A-Count (left); V-Count (right).
  \label{a_v_c}
  \vspace{0.2cm}
\end{minipage}
\begin{minipage}[b]{0.5\linewidth}
  \centering
  \centerline{\includegraphics[width=7.9cm]{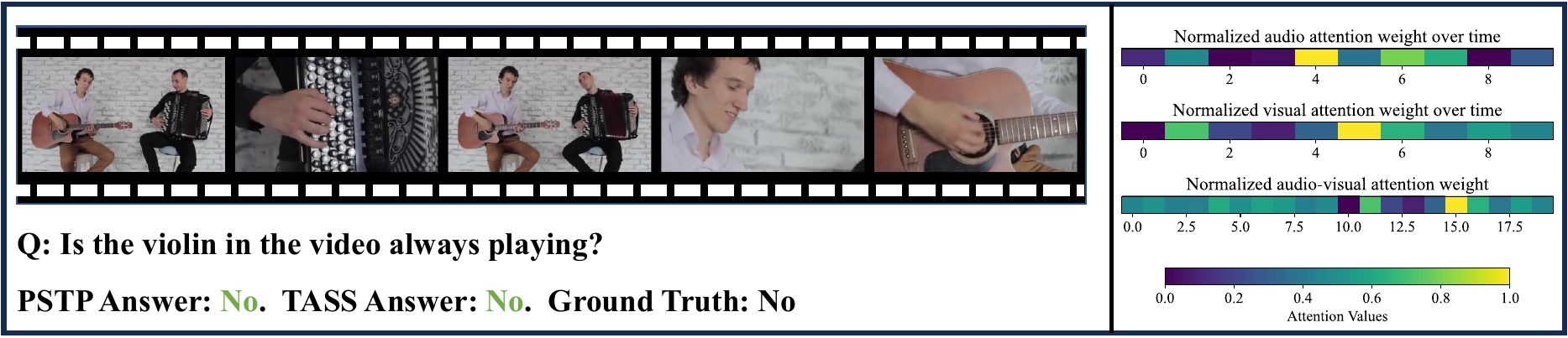}}
  (e) AV-Existential.
  \label{av_e}
  % \vspace{0.3cm}
\end{minipage}
\begin{minipage}[b]{0.5\linewidth}
  \centering
  \centerline{\includegraphics[width=7.9cm]{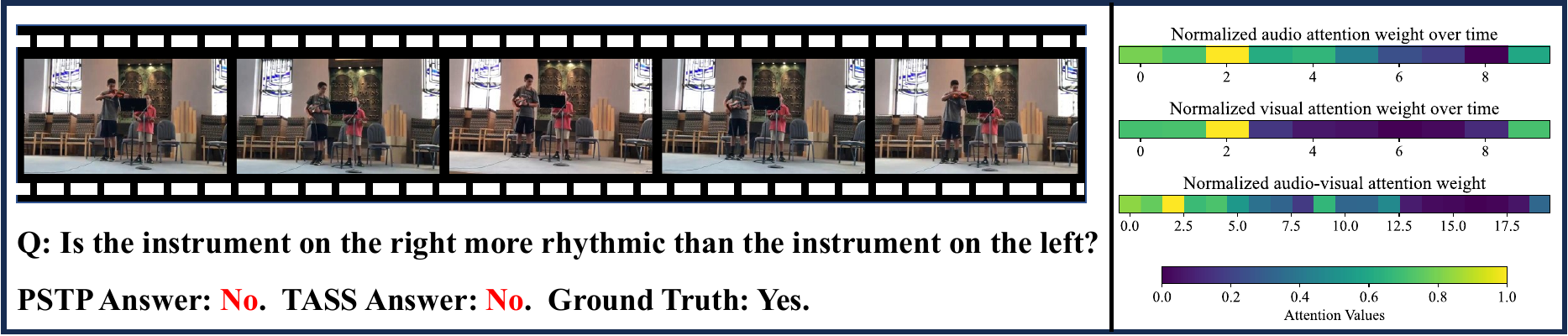}}
  (f) AV-Comparative.
  \label{av_C}
  % \vspace{0.3cm}
\end{minipage}
\caption{Qualitative results comparison against ground truth, and the visualization of attention weights on audio and visual segment features in our single-stream network: the first line is the normalized audio attention weights $\boldsymbol{w}_a$, the second line is the normalized visual attention weights $\boldsymbol{w}_v$, and the third line is the normalized audio-visual attention weights $[\boldsymbol{w}_a;\boldsymbol{w}_v]$ consists of $\boldsymbol{w}_a$ and $\boldsymbol{w}_v$ in order. The value of attention weights represents the correlation between each unimodal segment and the question, with the lighter the color, the greater the correlation. Then, our model aggregates the information of all timestamps based on the attention weights to predict the answer. The subfigure(f) shows one failure case predicted by our method.}
\label{fig5}
\end{figure*}

\textbf{Results Comparison.} In Fig. \ref{fig5}, we demonstrate the results of our method and conduct a comparison with state-of-the-art CLIP-based PSTP \cite{li2023progressive} method and the ground truth over 7 multi-modal scenes, namely audio-visual temporal, audio-visual location, audio-visual count, audio count, visual count, audio-visual existential and audio-visual comparative. The qualitative evaluation in Fig. \ref{fig5}(a-e) shows the ability of our method to perform better than the state-of-the-art in fine-grained (d.visual counting) and audio-involved questions (c.audio-visual location). As shown in the right example of Fig. \ref{fig5}(d), to answer the visual counting question, our TASS method predicts the correct answer `4' when PSTP \cite{li2023progressive} predicts wrong `6', which further demonstrates the effectiveness our proposed TSG+ module that transfers CLIP's knowledge of image-text matching to finer-grained region-text matching. In Fig. \ref{fig5}(c), PSTP predicts the wrong answer `Piano' and our method predicts the right answer `Trumpet'. As shown in Fig. \ref{fig5}(c), there is a prominent piano in the middle of visual stream, and PSTP used CLIP solely as a feature extractor, causing the audio features to be isolated by the aligned visual and text features. Consequently, even though only the trumpets sound in the video, PSTP predicts the wrong answer `Piano' due to the greater similarity between the text features and the image features. Differently, our CLIP-powered TASS approach utilizes the pretrained knowledge of CLIP to predict the correct answer by extending its image-text matching knowledge to audio-text matching. These results demonstrate the ability of our method to capture the sounding objects in key timestamps and have a comprehensive question-oriented understanding of the whole video.

\textbf{Single-stream Network.} In each of the examples in Fig. \ref{fig5}, we also visualize the normalized attention weights in our single-stream joint temporal grounding module, presented in three different ways: audio (intramodal), visual (intramodal), and audio-visual (multimodal). Note that audio and video are interleaved by segments in our single-stream network, but to visually show the proportion of each modality in answering different questions, we put the attention weights in audio-visual order. In the case of Fig. \ref{fig5}(a), to answer what instrument comes in first, we need to simultaneously hear the sound and see it in visual input, which requires temporal alignment between audio and visual information. Moreover, according to the temporal keyword: \textit{first}, the model should focus on the segment from the beginning of the video. As shown in the right figure of Fig. \ref{fig5}(a), attributed to our cross-modal synchrony loss during training, the attention weights in our model for the audio modality and the visual modality are temporally consistent, both focusing more on the beginning of the video, with the first segment having the largest attention weight, consistent with keyword \textit{first} in the question. Combined with the TSG+ module, our method successfully learns the spatial and temporal information and predicts the right answer \textit{violin} on the semantic level. 

In addition, based on the results of the visualization of the attention weights in different scenarios, we can obtain two observations related to our cross-modal synchrony loss and single-stream structure respectively. 

1) \textit{Soft synchronisation.} Although the attention weights on audio and visual segment features have similar temporal distributions (a-f), \textit{i.e.}, synchronously highlighting the front or middle part of the video, the attention weights for different modalities are not strictly the same, which can flexibly deal with possible audio-visual asynchrony in the video, as shown in Fig. \ref{fig5}(d) left and (e). 
% This enhances the model's ability to utilize the complex content in dynamic long-term audio-visual scenarios. 
This is attributed to our CMS loss, which uses JS divergence to constrain the distribution consistency of the attention weight between the question on the audio and visual features in the time dimension, using the question as a medium to maintain a soft temporal correlation between the audio-visual modalities, thereby enhancing the model’s ability to exploit complex information in dynamic audio-visual scenes.

2) \textit{Modality-aware.} According to the visualization results of the normalized audio-visual attention weights, we can see that our single-stream network pays more attention to the corresponding modal features when answering questions in unimodal scenes, as shown in Fig. \ref{fig5}(d). In the case of Fig. \ref{fig5}(d) right, to answer ``How many flute are in the entire video?'' in the visual counting scenario, our model assigns greater weight to the visual features, \textit{i.e.}, the second half of the color bar in the third row is lighter compared to the first half. The same is observed in the audio scene, as shown in Fig. \ref{fig5}(d) left. Moreover, when answering questions in audio-visual scenarios, our model has a different emphasis on different modalities for different questions. For example, as shown in Fig. \ref{fig5}(b), the model relies more on audio information when answering how many types of musical instruments there are (left), while the model relies more on visual information when answering how many flutes (of the same type) there are (right). This is consistent with human intuition because when multiple identical instruments are playing together, the number of instruments can be better counted using visual information as opposed to audio information. This interesting phenomenon is due to our single-stream network that seamlessly performs the audio-visual feature selection and fusion processes simultaneously for the first time.

\textbf{Failure Case.} We show failure case in Fig. \ref{fig5}(f) to analyse the result further. The audio-visual comparative question falls on the comparison of sound information, including comparison of rhythm or loudness. As shown in Fig. \ref{fig5}(f), while our method correctly aware the significance of audio modality in answering the question, it gives the wrong answer to whether the instrument on the right is more rhythmic than the instrument on the left. Insufficient exploration of audio features is the main reason for failure of Q\&A. In addition, the text features extracted by CLIP \cite{clip} are closer to the visual modality. As shown in Table \ref{tab1}, the accuracy of CLIP-based PSTP \cite{li2023progressive} method in answering audio questions is only 70.91\%. Nonethless, our method, which fully considers the natural audio-visual matching characteristics, extends CLIP's image-text matching knowledge to audio-text matching, surpassing PSTP by 5.01\% and reaching 75.92\%, but the accuracy of answering audio comparative questions is still unsatisfactory. We believe that this phenomenon can be improved by replacing the audio feature extractor with a more advanced one. 
% In addition, the reference of objects in this type of question involves orientation (on the left/right), which poses a challenge to CLIP, as it is based on semantic image-level contrastive learning.

\section{Conclusions}
This paper proposes a CLIP-powered target-aware single-stream network to better solve the question-oriented audio-visual scene understanding within the AVQA task. We provide novel approaches for exploiting the pretrained image-text matching knowledge and the natural audio-visual matching characteristic. The target-aware spatial grounding module exploits the explicit semantics of the question by leveraging the prior knowledge from the pretrained vision-language model, enabling the model to focus on the query subjects when parsing the audio-visual scenes. Also, the single-stream joint temporal grounding module treats audio and video as a whole through a single-stream structure and encourages the temporal association between audio and video with the proposed cross-modal synchrony loss. Extensive experiments have verified the superiority and robustness of the proposed module. Our work offers an inspiring new direction for audio-visual scene understanding and spatio-temporal reasoning in question answering.

\backmatter

\bmhead{Acknowledgements}

This work was supported partly by the National Natural Science Foundation of China (Grant No. 62173045, 62273054), partly by the Fundamental Research Funds for the Central Universities (Grant No. 2020XD-A04-3), and the Natural Science Foundation of Hainan Province (Grant No. 622RC675).

%%===========================================================================================%%
%% If you are submitting to one of the Nature Portfolio journals, using the eJP submission   %%
%% system, please include the references within the manuscript file itself. You may do this  %%
%% by copying the reference list from your .bbl file, paste it into the main manuscript .tex %%
%% file, and delete the associated \verb+\bibliography+ commands.                            %%
%%===========================================================================================%%
\bibliography{reference}
% \bibliography{sn-bibliography}% common bib file
%% if required, the content of .bbl file can be included here once bbl is generated
%%\input sn-article.bbl

\end{document}